\documentclass[11pt]{amsart}
\usepackage{amsbsy,amssymb,amscd,amsfonts,latexsym,amstext,delarray,
amsmath,graphicx} 
\input xypic

\usepackage{color}

\def\F{{\mathbb F}}

\def\Q{{\mathbb Q}}
\def\R{{\mathbb R}}

\def\cC{{\mathcal C}}

\def\cL{{\mathcal L}}

\def\cR{{\mathcal R}}

\title{Syntactic Structures and Code Parameters}
\author{Kevin Shu and Matilde Marcolli}
\address{Division of Physics, Mathematics, and Astronomy \\ California Institute of Technology \\
1200 E. California Blvd, Pasadena, CA 91125, USA}

\email{kshu@caltech.edu}
\email{matilde@caltech.edu}

\begin{document}
\maketitle

\begin{abstract}
We assign binary and ternary error-correcting codes to the data of syntactic
structures of world languages and we study the distribution of code points
in the space of code parameters. We show that, while most codes populate the
lower region approximating a superposition of Thomae functions, there is
a substantial presence of codes above the Gilbert--Varshamov bound and 
even above the asymptotic bound and the Plotkin bound. We investigate 
the dynamics induced on the space of code parameters by spin glass
models of language change, and show that, in the presence of entailment
relations between syntactic parameters the dynamics can sometimes improve
the code. For large sets of languages and syntactic data, one can gain
information on the spin glass dynamics from the induced dynamics in the
space of code parameters.
\end{abstract}

\section{Introduction}

This is a companion paper to \cite{Mar}, where techniques from coding theory were
proposed as a way to address quantitatively the distribution of syntactic features across
a set (or family) of languages. In this paper we perform a computational analysis, based
on the syntactic structures of world languages recorded in the SSWL database, and
on data of syntactic parameters collected by Longobardi and collaborators. We first
analyze all codes obtained from pairs and triples of languages in the SSWL database
and we show that their code points tend to populate the lower region of the
space of code parameters, approximating two scaled copies of the Thomae function.
Points that fall in that region behave essentially like random codes.
We then analyze arbitrary subsets of languages and syntactic data from the SSWL database 
and we compute the density of the distribution of their code points, showing that,
while most of them populate the lower region, there is a significant presence of
codes above the Gilbert--Varshamov bound and even above the asymptotic bound
and the Plotkin bound. We consider spin glass models of language change and
the induced dynamics on the space of code parameters, and show that, in the
presence of entailment relations the dynamics can enter the region above the 
Gilbert--Varshamov bound. We also show how the induced dynamics in the
space of code parameters can be helpful in gaining information on the behavior
of the spin glass model on large datasets of languages and parameters, where
convergence becomes very slow and a direct analysis of the dynamics 
becomes computationally difficult. 

\section{Codes and code parameters from syntactic structures}

In \cite{Mar} a new coding theory approach to measuring entropy and complexity
of a set of natural languages was proposed. The idea is to associate to each language
in the set a vector of binary variables that describe syntactic properties of the language.
The notion of encoding syntactic structures through a set of binary syntactic parameters
is a crucial part of the Principles and Parameters program of Linguistics developed
by Chomsky, \cite{Chomsky}, \cite{ChoLa}, see also \cite{Baker} for an expository
introduction. Thus, given a set of languages $\cL=\{ \ell_1, \ldots, \ell_N \}$ and a 
set of $n$ binary syntactic variables, whose values are known for all the $N$ languages,
one obtains a code $\cC_\cL$ in $\F_2^n$ consisting of $N$ code words. 

\smallskip

As argued in \cite{Mar}, one can use the properties of the resulting codes $\cC_\cL$, 
as an error correcting code, and its position in the space of code parameters to measure 
how syntactic features are distributed across the languages in the set. Moreover, the
position of the code $\cC_\cL$ in the space of code parameters, with respect to curves such as the
Gilbert--Varshamov bound and the asymptotic bound provide a measure of entropy and
of complexity of the set of languages $\cL$, which differs from measures of entropy/complexity
for an individual language. 

\smallskip
\subsection{Code parameters and bounds}

In the theory of error-correcting codes (see for instance \cite{TsfaVla}),  to a given 
code $\cC\subset \F_q^n$, one assigns two {\em code parameters}: the {\em transmission
rate}, or {\em relative rate} of the code, which measures how good the encoding procedure is, and which
is given by the ratio
$$ R(\cC)= \frac{k}{n}, \ \ \  \text{ with } \ \ k = \log_2(\# \cC)=\log_2(N), $$
where $k=\log_2(\# \cC)$ is the {\em absolute rate} of $\cC$, 
and the {\em relative minimum distance} of the code, which measures how good
the decoding is, and which is given by the ratio
$$ \delta(\cC) =\frac{d}{n}, \ \ \ \text{ with } \ \  d = \min_{\ell_1\neq \ell_2 \in C} d_H(\ell_1,\ell_2), $$
where $d_H(\ell_1,\ell_2)$ denotes the Hamming distance between the binary strings that
constitute the code words of $\ell_1$ and $\ell_2$,
$$ d_H(\ell_1,\ell_2) = \sum_{i=1}^n |x_i - y_i|,$$ for $\ell_1=(x_i)_{i=1}^n$
and $\ell_2=(y_i)_{i=1}^n$ in $\cC$, with $x_i, y_i \in \{ 0,1 \}$. 
Codes $\cC$ that have both $R(\cC)$ and $\delta(\cC)$ as large as possible are optimal
for error-correction, as they have simpler encoding and less error-prone decoding. In general,
it is not possible to arbitrarily improve both parameters, hence the quality of a code $\cC$ is 
estimated by the position of its code point $(\delta(\cC),R(\cC))$ in the space of code
parameters of coordinates $(\delta,R)$ inside the square $[0,1]\times [0,1]$. 
Various bounds on code parameters have been studied,
\cite{Man}, \cite{TsfaVla}, \cite{VlaDri}. 

\smallskip

As discussed in \cite{Mar} there are two bounds, that is, two curves in the space of
code parameters, that have an especially interesting meaning: the Gilbert--Varshamov
curve, which is related to the statistical behavior of random codes (see \cite{BaFo}, \cite{CoGo}),
and the asymptotic bound, whose existence was proved in \cite{Man}, further studied in
\cite{Man2}, \cite{Man-CE}, \cite{ManMar}, \cite{ManMar2}. The asymptotic bound separates
the region where code points are dense and have infinite multiplicity from the region where
they are sparse and with finite multiplicity, \cite{Man2}, \cite{ManMar2}.
The Gilbert--Varshamov curve for
a $q$-ary code has a simple form, 
$$ R=1-H_q(\delta), \ \ \ \text{ with } \ \  H_q(\delta)=\delta\, \log_q(q-1) - \delta\, \log_q\delta 
- (1-\delta)\,\log_q (1-\delta), $$
the $q$-ary Shannon entropy.  With the asymptotic bound, however, the situation is much
more complicated, as one does not have an explicit expression. Indeed, the question of
the computability of the asymptotic bound was posed in \cite{Man2}, and addressed in
\cite{ManMar2} in terms of a relation to Kolmogorov complexity (which is not a computable function). 
Namely, it is shown in \cite{ManMar2} that the asymptotic bound becomes computable, given
an oracle that can order codes by increasing Kolmogorov complexity.  Even though one does
not have an explicit expression for the asymptotic bound, several estimates on its location in
the space of code paramaters are described in \cite{TsfaVla}. Thus, in practical cases, it will
be possible to obtain sufficient conditions to check if a code point violates the asymptotic bound
by using some of these estimates.  In particular, there is a relation between the asymptotic
curve $R=\alpha_q(\delta)$ and the Gilbert--Varshamov bound,
$$ \alpha_q(\delta) \geq 1 - H_q(\delta), $$
and a relation between the asymptotic bound and the Plotkin bound
$R \leq 1- \frac{\delta}{q}$. The Plotkin line lies above the Gilbert--Varshamov  curve
and the asymptotic bound satisfies $\alpha_q(\delta) \leq 1- \frac{\delta}{q}$, with
$\alpha_q(\delta)=0$ for $(q-1)/q < \delta \leq 1$.  We will be considering only binary
codes, hence we have everywhere $q=2$. 

\smallskip
\subsection{Bilingual and trilingual syntactic codes}

In this companion paper to \cite{Mar}, we carry out some analysis, based on the binary
syntactic variables recorded in the SSWL database ``Syntactic Structures of World Languages",
\cite{SSWL}. We will not refer to these variables as ``syntactic parameters", because it is
understood among linguists that the binary variables recorded in the SSWL database do not
correspond to ``syntactic parameters" the sense of the Principles and Parameters program,
for example because of conflation of deep and surface structure. However, this database is
useful because it contains a fairly large number of world languages (253) and of syntactic
variables (115). From a computational perspective, one of the main problems in using SSWL
data lies in the fact that not all 115 variables are mapped for all the 253 languages: indeed the
languages are very non-uniformly mapped, with some (mostly Indo-European) languages
mapped with 100$\%$ of the variables and others with only very few entries. Thus, for the
purpose of the computations in this paper, when comparing a set of different languages, we
have only used those variables that are fully mapped, in the SSWL database, for all the
languages in the given set. 

\smallskip

We consider here all the pairs $\{ \ell_1, \ell_2 \}$ of languages in the SSWL database,
and all triples $\{ \ell_1, \ell_2, \ell_3 \}$, and the resulting codes $\cC_{\ell_1,\ell_2}$
and $\cC_{\ell_1,\ell_2,\ell_3}$. We refer to these as ``bilingual and trilingual syntactic codes".
The computation grows rapidly much heavier for larger sets of languages. However, these 
cases are enough to see some interesting results on how the corresponding code
parameters are distributed in the space of code parameters. As explained above, the length 
of the code words for these codes is not fixed: it is the largest number of syntactic binary
variables in the SSWL list that are completely mapped for the languages in the set. Thus,
for a set $\{ \ell_1, \ell_2 \}$ or $\{ \ell_1, \ell_2, \ell_3 \}$ we have a corresponding number
of binary variables $n=n(\ell_1, \ell_2)$ or $n=n(\ell_1, \ell_2, \ell_3)$ with the codes
$\cC_{\ell_1,\ell_2}\subset \F_2^{n(\ell_1, \ell_2)}$ and $\cC_{\ell_1,\ell_2,\ell_3} \subset
\F_2^{n(\ell_1, \ell_2, \ell_3)}$. This is not a problem, since points in the space of
code parameters correspond to codes of any arbitrary length. For each code 
$\cC_{\ell_1,\ell_2}$ and $\cC_{\ell_1,\ell_2,\ell_3}$ we compute the corresponding
code parameters $$ (\delta(\cC_{\ell_1,\ell_2}),R(\cC_{\ell_1,\ell_2})) \ \ \text{ and } \ \ 
(\delta(\cC_{\ell_1,\ell_2, \ell_3}),R(\cC_{\ell_1,\ell_2, \ell_3})), $$
and we plot them in the plane of code parameters. We then compare their
position to different bounds in the space of code parameters. 

\begin{figure}[h]
    \centering
    \includegraphics[width=0.9\linewidth]{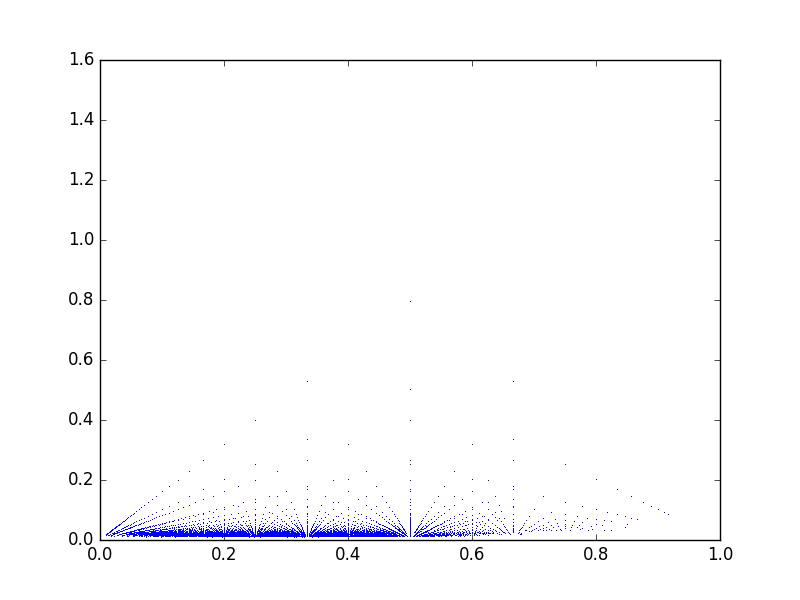}
    \caption{The code parameters of all 2 and 3 tuples of languages in the SSWL database.}
    \label{fig:code_params1}
\end{figure}

\smallskip
\section{Code parameters of syntactic codes}

The Python code is available at {\tt www.its.caltech.edu/$\sim$matilde/SSWLcodes}.
The {\tt code\_params.py} script has utilities that will compute the code parameters for 
a given subset of the languages. The plot for all the codes $\cC_{\ell_1,\ell_2}$
and $\cC_{\ell_1,\ell_2,\ell_3}$ of pairs and triples of languages in the SSWL database
was generated with {\tt fixed\_sized\_subset.py}. The resulting plot is given in 
Figure~\ref{fig:code_params1}. Note that for all the codes in this set the absolute rate 
of the code is always either $k(\cC_{\ell_1,\ell_2})=1$ or 
$k(\cC_{\ell_1,\ell_2,\ell_3})=\log_2(3)$.

\smallskip

\subsection{Thomae function and random parameters}

The fractal pattern that one sees appearing in Figure~\ref{fig:code_params1} may seem an
first surprising, but in fact it has a very simple explanation. Recall that the Thomae function
defined as
$$ f(x) = \left\{ \begin{array}{ll} 1 & x=0 \\ 
\frac{1}{q} & x\in \Q,\, x=\frac{p}{q}, \, q>0 \\
0 & x\in \R \smallsetminus \Q \end{array} \right. $$
\begin{center}
\begin{figure}
\includegraphics[width=0.7\linewidth]{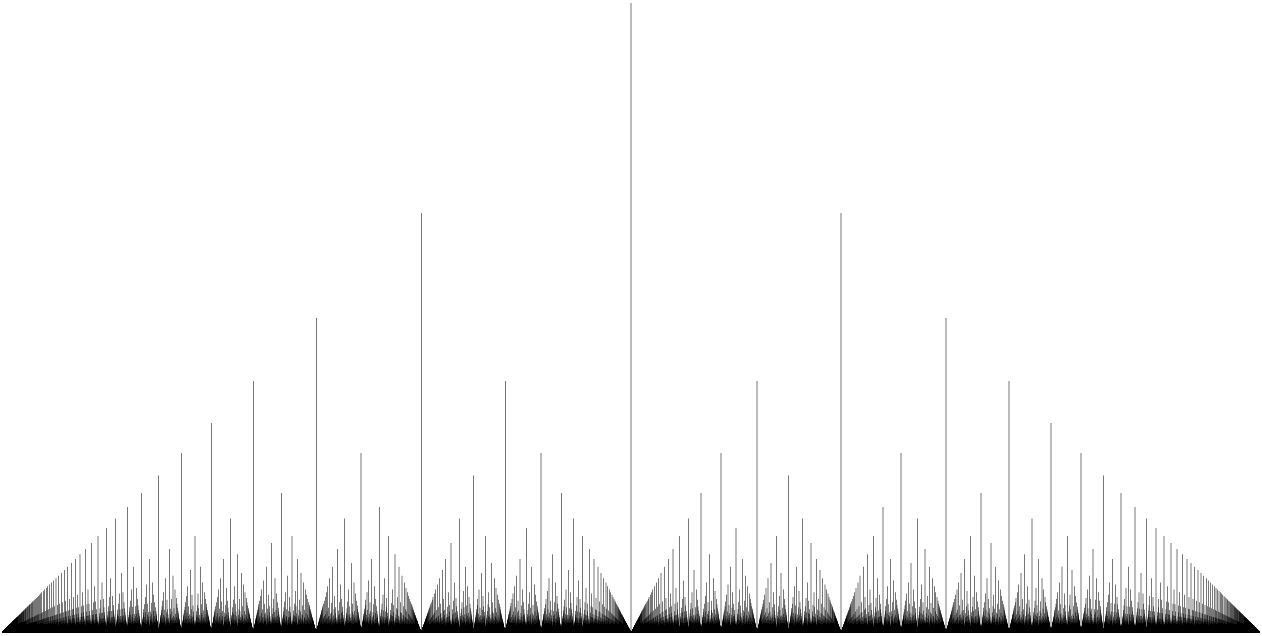}
\caption{The Thomae function. \label{Thomfig}}
\end{figure}
\end{center}
A plot of the undergraph of the Thomae function is shown in Figure~\ref{Thomfig}. 
One can clearly see the similarity with Figure~\ref{fig:code_params1}. Indeed, the
reason why the code parameters in Figure~\ref{fig:code_params1} approximate the
Thomae function depends on the fact that we are fixing the absolute rate of the codes.
We have code points
$$ (\delta(\cC), R(\cC)) = (\frac{d}{n}, k \cdot \frac{1}{n} ) . $$
Thus, since $k$ is fixed to be either $1$ or $\log_2(3)$, we obtain two copies, scaled by
the respective values of $k$, of the graph of
$$ f: \frac{d}{n} \mapsto \frac{1}{n}, $$
which, when $(d,n)=1$, agrees with the Thomae function, restricted to those values of $d$ and $n$ that occur
in our set of codes. We see that we obtain several additional points in Figure~\ref{fig:code_params1} that 
lie in the undergraph of the Thomae function, which come from the cases with $(d,n)\neq 1$, where 
for $d=u r$ and $n= v r$, in addition to the plot point $(\frac{u}{v} ,\frac{1}{v})$ on the graph of the 
Thomae function, we also find other lower plot points $(\frac{u}{v}, \frac{1}{vr})$ on the same vertical line. 

\smallskip

\begin{center}
\begin{figure}
    \includegraphics[width=0.8\linewidth]{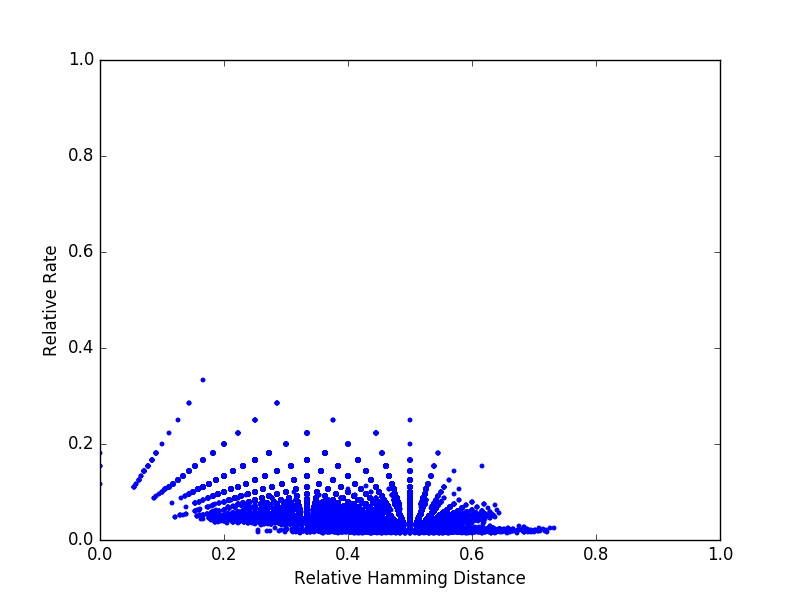}
    \caption{Plot for $3$-tuples with randomly generated parameters. \label{fig:random_params}}
\end{figure}    
\end{center}

We compare this behavior with the case of code parameters produced by randomly generated points.
A plot of $3$-tuples of a set of randomly generated parameters is shown in Figure~\ref{fig:random_params}. 
This was generated with {\tt random\_parameters.py}. We see that the lower region of 
Figure~\ref{fig:code_params1} behaves similarly to the case of randomly generated parameters.
Indeed, the region in the space of code parameters that lies below the Gilbert--Varshamov line is typically
the one that is populated by random codes, \cite{BaFo}, \cite{CoGo}. 

\smallskip
\subsection{Gilbert--Varshamov and Plotkin bound}

We want to analyze the position, with respect to various bounds, of the code parameters 
of codes $\cC_\cL$, for sets $\cL$ of languages and their SSWL syntactic data. In order to
do that, since systematic computations for larger sets of languages become lengthy, we
check the position of code points on sets of randomly selected languages and syntactic
parameters in the SSWL database. 
The script {\tt random\_subset.py} takes the list of binary syntactic variables 
recorded in the SSWL database and selects a random choice of a subset of
these binary variables. Then it randomly selects a subset of languages, 
among those for which the selected set of parameters is completely mapped in
the database. 

\begin{center}
\begin{figure}
    \includegraphics[width=0.9\linewidth]{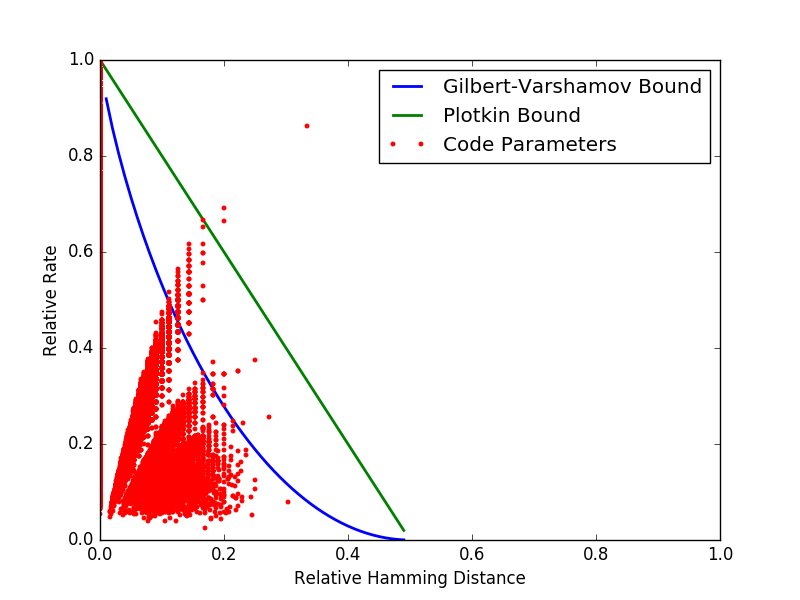}
    \caption{Code points for set of randomly selected SSWL syntactic variables and languages next to the Gilbert--Varshamov and the Plotkin bounds. \label{fig:random_subset_w_bounds}}
\end{figure}
\end{center}

\begin{center}
\begin{figure}
    \includegraphics[width=0.9\linewidth]{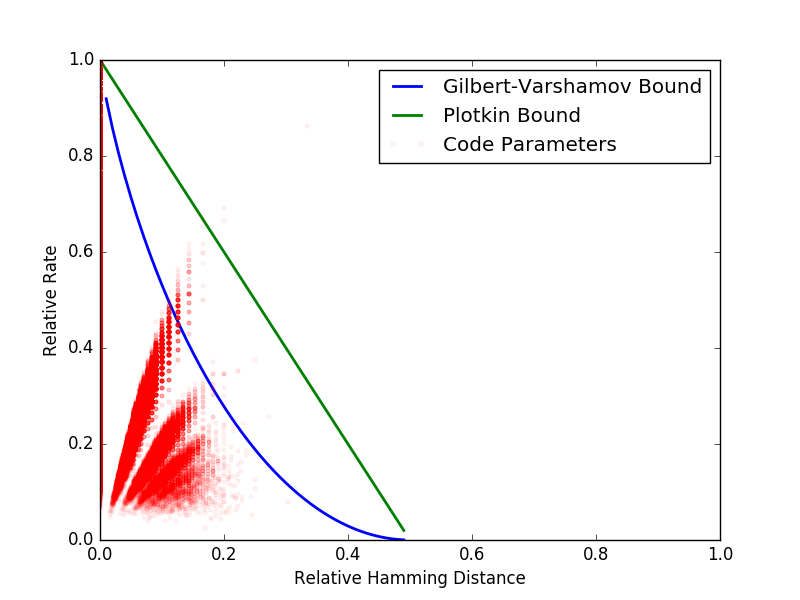}
    \caption{Density of code points of syntactic codes, with respect to the Gilbert--Varshamov and the Plotkin bounds. \label{fig:sub2}}
\end{figure}
\end{center}

In a typical plot obtained with this method, like the one shown in Figure~\ref{fig:random_subset_w_bounds},
we see curves that correspond to a superposition of several scaled Thomae functions, for varying values
of $k$. A large part of the code points tends to cluster in the region below the
Gilbert--Varshamov curve, similarly to what one would expect for random codes. However, there
is typically a significant portion of the code points obtained in this way that lies above the 
Gilbert--Varshamov, populating the area between the Gilbert--Varshamov curve and the Plotkin line.
Since these points tend to cover the entire width of the region between these two curves, which contains
the asymptotic bound, certainly part of them will lie above the asymptotic bound, in the region of the
sporadic codes. One usually sees an even smaller number of code points that lie above the Plotkin bound
(hence certainly above the asymptotic bound).  This confirms the observation made in \cite{Mar}
regarding code points of syntactic codes and their position in the space of code parameters. 
Using randomized sets of SSWL parameters and corresponding sets of languages, we can also
plot the density of code points in the various regions of the space of code parameters, as shown in
Figure~\ref{fig:sub2}.

\smallskip
\section{Dynamics in the space of code parameters}

In \cite{STM} a dynamical model of language change was proposed, based on 
a spin glass model for syntactic parameters and language interactions. It is a simple
model based on a graph with languages at the vertices, represented by a vector 
of their syntactic parameters interpreted as spin variables, and strengths of interaction 
between languages along the edges, measured using data proportional to the amount 
of bilingualism. In the case of syntactic parameters behaving as independent variables,
in the low temperature regime (see \cite{STM} for a discussion of the interpretation 
of the temperature parameter in this model) the dynamics converges rapidly towards
an equilibrium state where all the spin variables corresponding to a given syntactic
feature for the various languages align to the value most prevalent in the initial
configuration. The SSWL database does not record relations between parameters,
although it can be shown by other approaches that interesting relations are present,
see \cite{Park}, \cite{Port}. 
Using syntactic data from \cite{Longo1}, \cite{Longo2}, which record explicit entailment
relation between different parameter, it was shown in \cite{STM}, for small graph
examples, that in the presence of relations the dynamics settles on equilibrium states
that are not necessarily given by completely aligned spins.

\smallskip
\subsection{Spin glass models for syntactic parameters}

When we interpret the dynamics of the model considered in \cite{STM} in terms
of codes and the space of code parameters, the initial datum of the set of languages $\cL$
at the vertices of the graph, with its given list of syntactic binary variables, determines
a code $\cC_\cL$. The absolute rate $k=k(\cC_\cL)= \log_2(\# \cL)$ and the number
of syntactic features considered $n=n(\cC_\cL)$ remain fixed along the dynamics, hence
the dynamics moves the code points along the horizontal lines with fixed $R$-coordinate. 
In the case of independent syntactic binary variables, the dynamics follows a gradient
descent for an energy functional that is simply given by the Hamiltonian
$$ H_{x_i} =- \sum_{\ell \neq \ell' \in \cL} J_{\ell, \ell'}  S_{x_i(\ell)} S_{x_i(\ell')}, $$
for a given syntactic variable $x_i$, $i=1,\ldots,n$, where $J_{\ell, \ell'}$ is the
strength of the interaction along the edge connecting the vertices $\ell$ and $\ell'$
and $S_{x_i}$ is the $\pm 1$ values spin variable associated to the binary variable
$x_i$. The minimum of the energy $H_{x_i}$ for the single variable $x_i$ is
achieved when $S_{x_i(\ell)} S_{x_i(\ell')}=1$, that is, when $| x_i(\ell)-x_i(\ell')|=0$.
Thus, in this case where each syntactic variables runs as an independent Ising 
model, the minimum is achieved where $d_H(x(\ell),x(\ell'))=\sum_{i=1}^n |x_i(\ell)-x_i(\ell')|=0$,
that is, when all the spins align. In the presence of entailment relations between different
syntactic variables, it was shown in \cite{STM} that the Hamiltonian should be modified
by a term that introduces the relations as a Lagrange multiplier. This alters the dynamics
and the equilibrium state, depending on a parameter that measures how strongly enforced the
relations are. From the point of view of coding theory discussed here, it seems more 
reasonable to modify this dynamical system, so that it can be better described as
a dynamics in the space of code parameters. It is natural therefore to consider a
similar setting, where we assign a given set $\cL$ of languages to the vertices
of a complete graph $G_\cL$, with assigned energies $J_e =J_{\ell,\ell'}$ at the edges
$e\in E(G_\cL)$ with $\partial e=\{ \ell, \ell'\}$. We denote by $x(\ell)=(x_j(\ell))_{j=1}^n$
the vector of binary variables that lists the $n$ syntactic features of the language $\ell$.
We consider these as maps $x: \cL \to \{0,1\}^n$, or equivalently as points 
$x \in \{0,1\}^{n\, \cL}$. Consider an energy functional of the form
$$ H(x) = \sum_{\ell\neq \ell' \in \cL} J_{\ell,\ell'} \, d_H(x(\ell), x(\ell')), $$
for $J_{\ell,\ell'}=J_e >0$, where $d_H(x(\ell), x(\ell'))$ is the Hamming distance, 
$$ H(x) = \sum_{e \in E(G_\cL)} \sum_{j=1}^n J_e \, | x_j(\ell) - x_j(\ell') |. $$
The corresponding partition function is given by
$$ Z = \sum_{x\in \{0,1\}^{n\, \cL}} e^{-\beta H(x)}. $$
At low temperature (large $\beta$), the partition function is concentrated around the
minimum of $H(x)$, that is, were all $d_H(x(\ell), x(\ell'))=0$, hence where all the vectors
$x(\ell)\in \{0,1\}^n$ agree. Given an initial condition $x^0 \in \{0,1\}^{n\, \cL}$ and
the datum $( J_e )_{e\in E(G_\cL)}$ of the strengths of the interaction energies along
the edges, the same method used in \cite{STM}, based on the standard Metropolis--Hastings 
algorithm, can be used to study the dynamics in this setting, with a similar behavior. 
In the space of code parameters, given the code point $(\delta^0,R^0)=(\delta(\cC(x^0)), R(\cC(x^0)))$
associated to the initial condition $x^0$, the dynamics moves the code point along the line
with constant $R=R^0$. As the dynamics approaches the minimum of the action, the
code point enters the region below the Gilbert--Varshamov bound, as it moves towards smaller
values of $\delta$. 

\smallskip
\subsection{Dynamics in the presence of entailment relations}

It is more interesting to see what happens in the case of where the syntactic variables are
not independent but involve entailment relations between syntactic parameters. To this
purpose we need to use syntactic data from \cite{Longo1}, \cite{Longo2}, where relations
between syntactic parameters are explicitly recorded. A typical form of the relations
described in  \cite{Longo1}, \cite{Longo2} consists of two syntactic parameters $x_i$ and $x_j$,
where $x_i$ is unconstrained at has binary values $x_i\in \{ 0,1 \}$, while the values
of $x_j$ are constrained by the value of $x_i$, so that if $x_i=1$ $x_j$ can take any
of two binary values, while if $x_i=0$ then $x_j$ becomes undefined. We express this
by considering $x_j$ as a ternary valued variables, $x_j\in \{ -1,0,+1 \}\simeq \F_3$,
where $x_j = \pm 1$ stand for the ordinary binary values and $x_j=0$ signifies undefined.
We can then write the relation in the form of a function 
$$ \cR_{ij}(x)=| x_i  - |x_j||, $$
where the solutions to $\cR_{ij}(x)=0$ are precisely $(x_i=1, x_j=\pm 1)$ and $(x_i=0,x_j=0)$.
We introduce a parameter $E_{ij}\geq 0$ that measures how strongly the relation $\cR_{ij}$
is enforced. The modified energy functional that accounts for the presence of a relation
between the $i$-the and the $j$-th parameter is then of the form
$$ H(x) = \sum_{\ell\neq \ell' \in \cL} J_{\ell,\ell'} \, d_H(x(\ell), x(\ell')) + \sum_\ell E_{ij}(\ell) \, \cR_{ij}(x(\ell)), $$
where we allow here for the possibility that the relation may be differently (more strongly or weakly)
enforces for different $\ell\in \cL$. In practice, it will be convenient to assume that $E_{ij}(\ell)=E_{ij}$
is independent of $\ell\in \cL$. When considering all the possible relations between different parameters
in the list, that are of the form $\cR_{ij}$ as above, we separate out the set
$\{1,\ldots, n\}$ of all the syntactic parameters in the list in two sets, $\{1,\ldots, n\}=B \cup T$,
where $B$ is the set of independent binary variables and $T$ is the set of entailed ternary variables,
and we write the energy functional as 
$$ H(x) = \sum_{\ell\neq \ell' \in \cL} J_{\ell,\ell'} \, d_H(x(\ell), x(\ell')) + \sum_{i\in B, j\in T} E_{ij} \, \cR_{ij}(x), $$
where 
$$ \cR_{ij}(x) =\sum_{\ell\in \cL} \cR_{ij}(x(\ell)) $$
and where $E_{ij}=0$ if there is no direct dependence of $x_j$ upon $x_i$. One can consider additional
terms $\cR_{i_1,\ldots, i_r. j}(x)$ with $i_a\in B$, $j\in T$ and $j\in T$ of a similar form, when the ternary parameter
$x_j$ is entailed by more than one binary parameter $x_i$, and add them to the energy functional in a similar way,
with entailment energies $E_{i_1,\ldots, i_r, j}\geq 0$.

\begin{center}
\begin{figure}
    \includegraphics[width=0.6\linewidth]{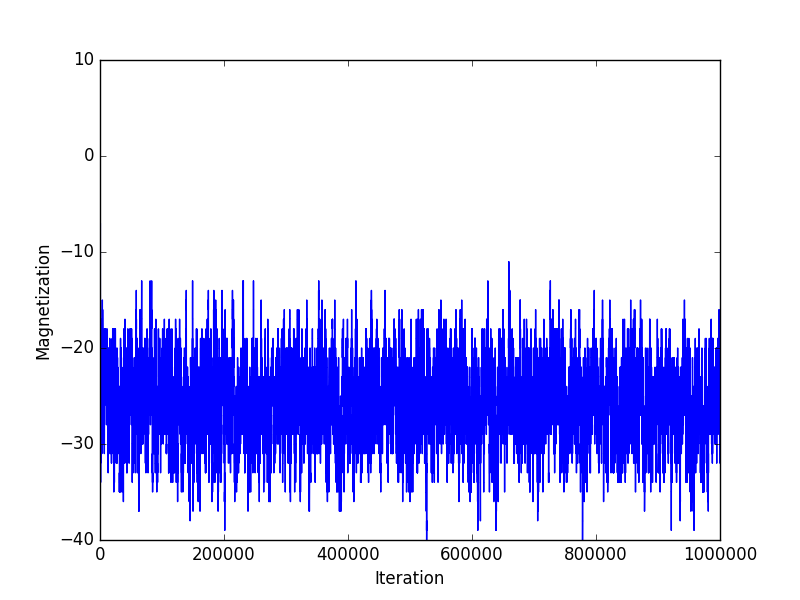}
    \includegraphics[width=0.6\linewidth]{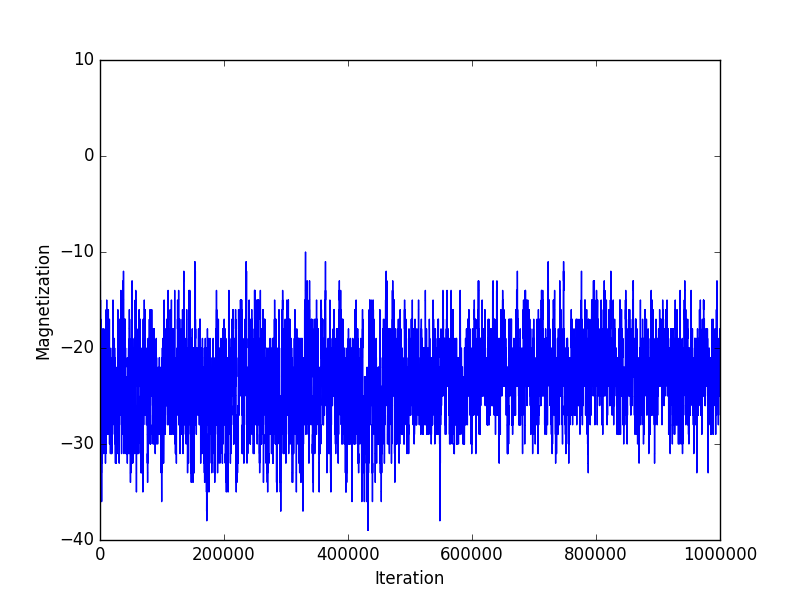}
    \includegraphics[width=0.6\linewidth]{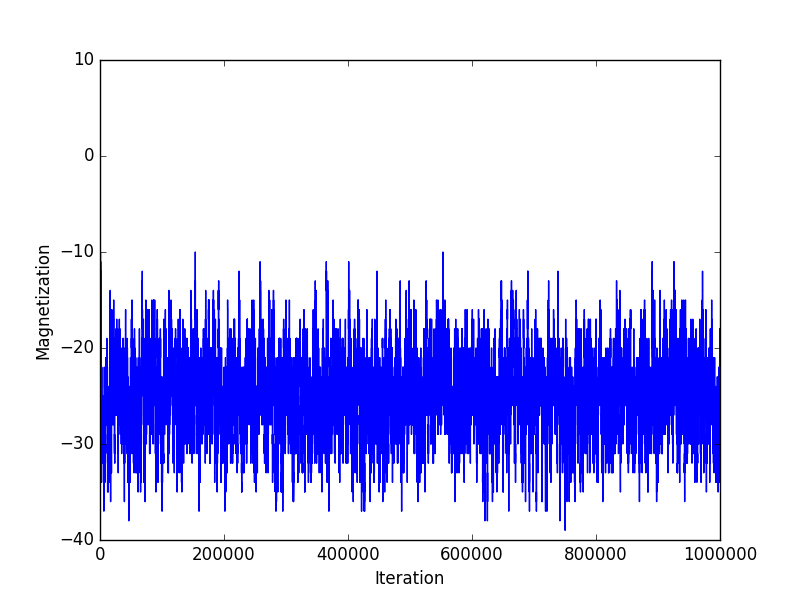}
    \caption{Average magnetization for the spin glass model of \cite{STM} computed for the 
    languages and parameters of \cite{Longo1}, in the cases with  $T=10$ and $E=0$; 
    $T=10$ and $E=9000$; $T=910$ and $E=0$. \label{magFig}}
\end{figure}
\end{center}

\smallskip
\subsection{Dynamics in the space of code parameters}

In these models, where part of the syntactic variables $x_i \in B$ are seen as binary variables
and part $x_j\in T$ as ternary variables, for the purpose of coding theory, we consider the whole
$x=(x_i)_{i=1}^n$ as a vector in $\F_3^n$, in order to compute the code parameters of
the resulting code $\cC(\cL)\subset \F_3^n$.

\smallskip

One can see already in a very simple
example, and using the dynamical system in the form described in \cite{STM},  
that the dynamics in the space of code parameters now does not need to move
towards the $\delta=0$ line. Consider the very small example, with just two entailed
syntactic variables and four languages, discussed in \cite{STM}, where the chosen languages
are  $\cL= \{ \ell_1, \ell_2, \ell_3, \ell_4 \} = \{{\rm English,Welsh, Russian,Bulgarian} \}$ and the
two syntactic parameters are $\{ x_1, x_2 \} = \{{\rm Strong Deixis, Strong Anaphoricity} \}$.
Since we have an entailment relation, the possible values of the variables $x_i$ are now
ternary, $x_i(\ell)\in \{ 0, -1, +1 \}$, that is, we consider here codes $\cC\subset \F_3^n$.
In this example $n=2$. The initial condition $x^0$ is given by 
$$  \begin{array}{l} x^0(\ell_1)= (+1, +1) \\ x^0(\ell_2)= (-1, 0) \\ x^0(\ell_3)= (+1, +1) \\ x^0(\ell_4)= (+1, +1) .
\end{array} $$
Note that, since we have two identical code words $x^0(\ell_1)=x^0(\ell_3)$ in this initial condition,
the parameter $d(\cC_\cL)=0$, so the code point $(\delta(\cC_\cL),R(\cC_\cL))=(0,\log_3(2)))$
already lies on the vertical line $\delta=0$. We consider in this case the same dynamical system
used in \cite{STM} to model the case with entailment, which is a modification of the Ising model
to a coupling of an Ising and a Potts model with $q=3$ at the vertices of the graph. This
dynamics, which depends on the temperature parameter $T=1/\beta$ an on an auxiliary
parameter $E$, the ``entailment energy", that measures how strongly the entailment
relation is enforced.  In the cases with high
temperature and either high or low entailment energy, it is shown in \cite{STM} that one can have 
equilibrium states like
$$ \begin{array}{l} x(\ell_1)= (+1, 0) \\ x(\ell_2)= (+1, -1) \\ x(\ell_3)= (-1, 0) \\ x(\ell_4)= (+1, +1) ,
\end{array} $$
for the high entailment energy case, or
$$ \begin{array}{l} x(\ell_1)= (+1, -1) \\ x(\ell_2)= (-1, -1) \\ x(\ell_3)= (-1, +1) \\ x(\ell_4)= (-1, -1) ,
\end{array} $$
for the low entailment energy case. In both of these cases, the minimum distance
$d=\min_{\ell\neq \ell'} d_H(x(\ell),x(\ell'))=1$, hence $\delta = 1/2$. Thus, along the dynamics,
the code point in the space of code parameters has moved away from the line $\delta=0$, along
the line with constant $R$. The final code point with $\delta=1/2$ and $R=\log_3(2)$ lies above
the GV-curve $R=1-H_3(\delta)$. Thus, in this very simple example we have seen that the
dynamics in the case where syntactic parameters are not independent variables can in fact
move the code toward a better code, passing from below to above the GV-bound. 

\begin{center}
\begin{figure}
    \includegraphics[width=0.6\linewidth]{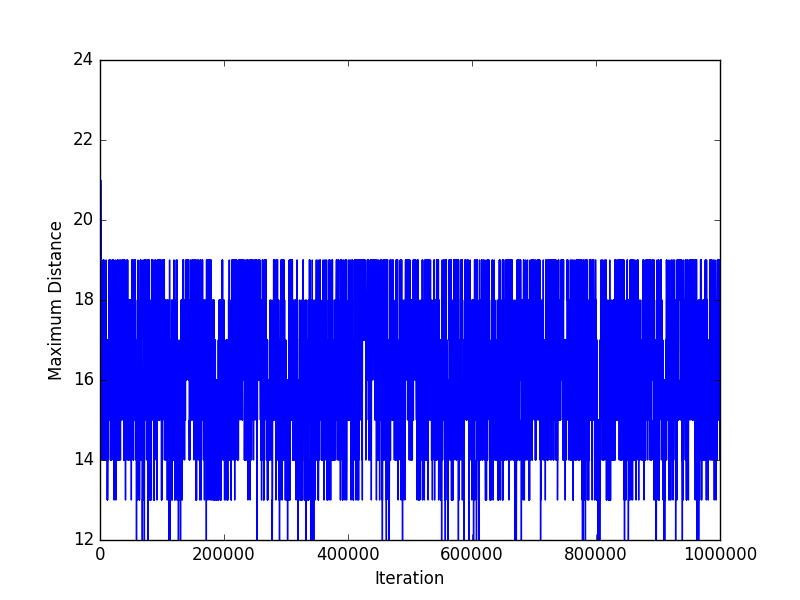}
    \includegraphics[width=0.6\linewidth]{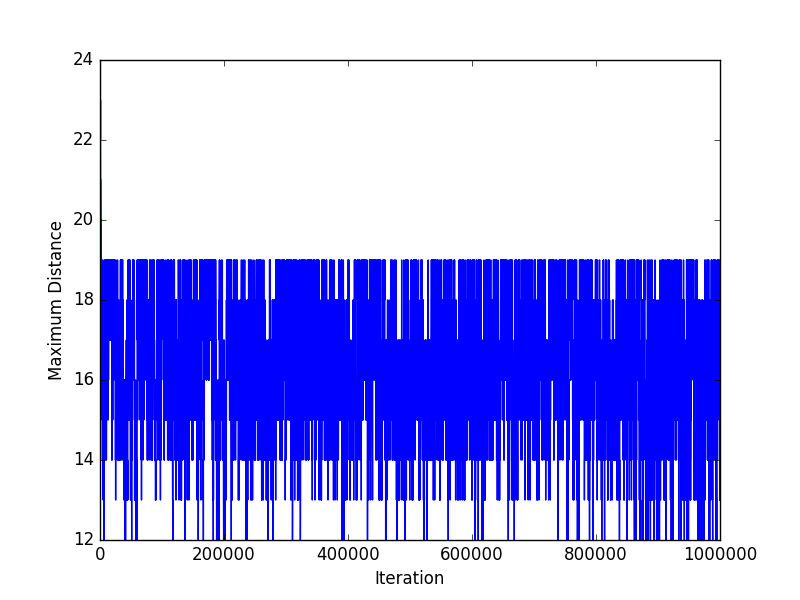}
    \caption{Dynamics in the space of code parameters: maximal distance. \label{distmaxFig}}
\end{figure}
\end{center}

\smallskip
\subsection{Simulations}

The example mentioned above is too simple and artificial to be significant, but we can analyze a more
general situation, where we consider the full syntactic data of \cite{Longo1}, \cite{Longo2},
with all the entailment relations taken into account, and the same interaction energies along
the edges as in \cite{STM}, taken from the data of \cite{Medialab}, which can be regarded as 
roughly proportional to a measure of the amount of bilingualism. 

\smallskip

\begin{center}
\begin{figure}
    \includegraphics[width=0.6\linewidth]{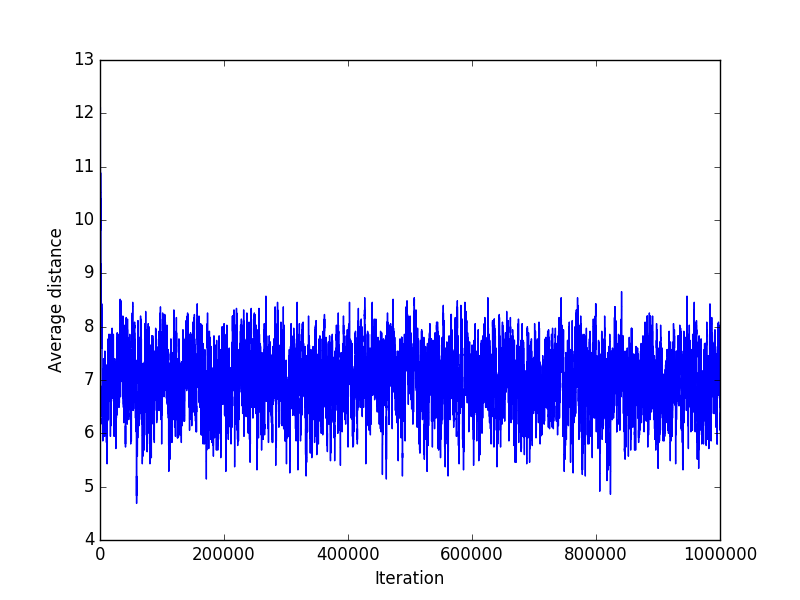}
    \includegraphics[width=0.6\linewidth]{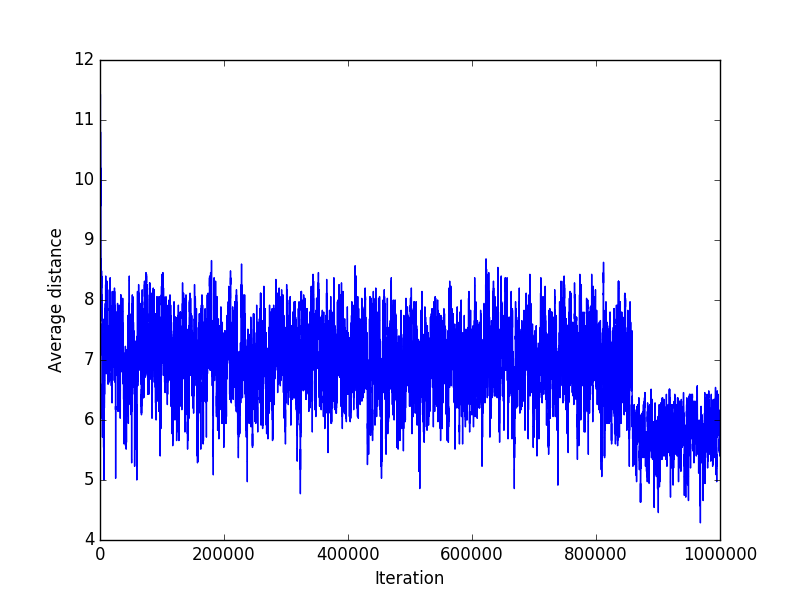}
    \caption{Dynamics in the space of code parameters: average distance. \label{distFig}}
\end{figure}
\end{center}

When we work with the full set of data from \cite{Longo1}, \cite{Longo2}, involving 63 parameters
for 28 languages (from which we exclude those that do not occur in the \cite{Medialab} data), 
we see that the large size of the graph and the presence of many entailment relations render
the dynamics a lot more complicated than the simple examples discussed in \cite{STM}. Indeed
for such a large graph the convergence of the dynamics becomes extremely slow, even in the
low temperature case and even when entailment relations are switched off, as shown in the graph of
the average magnetization in Figure~\ref{magFig}. Such a large system becomes computationally
too heavy, and it is difficult to handle a sufficiently large iterations to get to see any 
convergence effect. 
However, when one
considers codes obtained by extracting arbitrary subsets of three languages from this set and
follows them along the dynamics, computing the corresponding position in the space of code
parameters, one sees that, in the case without entailment ($E=0$) the average distance drops 
notably after enough iteration, as shown in Figure~\ref{distFig} indicating that the simulation 
might in fact converge, even
though at the same state in the number of iteration the average magnetization is not settling
yet. In the case with entailment relations one should expect the convergence process to be even
slower. Moreover, as in the small example discussed above, the $\delta$ parameter 
may settle on a limit value different than zero, so the data of the simulation are less informative.

\end{document}